# Open Information Extraction

Duc-Thuan Vo and Ebrahim Bagheri

Laboratory for Systems, Software and Semantics (LS³),
Ryerson University, Toronto, ON, Canada
{thuanvd, bagheri}@ryerson.ca

Open Information Extraction (Open IE) systems aim to obtain relation tuples with highly scalable extraction in portable across domain by identifying a variety of relation phrases and their arguments in arbitrary sentences. The first generation of Open IE learns linear chain models based on unlexicalized features such as Part-of-Speech (POS) or shallow tags to label the intermediate words between pair of potential arguments for identifying extractable relations. Open IE currently is developed in the second generation that is able to extract instances of the most frequently observed relation types such as Verb, Noun and Prep, Verb and Prep, and Infinitive with deep linguistic analysis. They expose simple yet principled ways in which verbs express relationships in linguistics such as verb phrase-based extraction or clause-based extraction. They obtain a significantly higher performance over previous systems in the first generation. In this paper, we describe an overview of two Open IE generations including strengths, weaknesses and application areas.

*Keywords*: Open Information Extraction, Natural Language Processing, Verb Phrase-based Extraction, Clause-based Extraction.

## 1. Information extraction and open information extraction

Information Extraction (IE) is growing as one of the active research areas in artificial intelligence for enabling computers to read and comprehend unstructured textual content (Etzioni et al., 2008). IE systems aim to distill semantic relations which present relevant segments of information on entities and relationships between them from large numbers of textual documents. The main objective of IE is to extract and represent information in a tuple of two entities and a relationship between them. For instance, given the sentence "Barack Obama is the President of the United States", they venture to extract the relation tuple PresidentOf (Barack Obama, the United States) automatically. The identified relations can be used for enhancing machine reading by building knowledge bases in Resource Description Framework (RDF) or ontology forms. Most IE systems [4, 14, 22, 27] focus on extracting tuples from domain-specific corpora and rely on some form of pattern-matching technique. Therefore, the performance of these systems is heavily dependent on considerable domain specific knowledge. Several methods employ advanced pattern matching techniques in order to extract relation tuples from knowledge bases by learning patterns based on labeled training examples that serve as initial seeds.

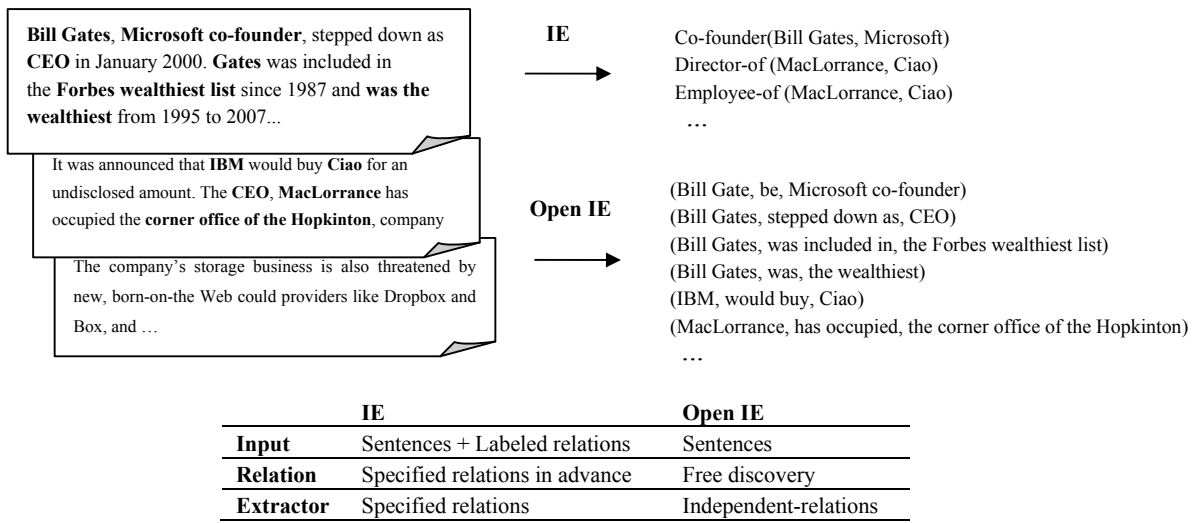

Fig. 1. IE vs. Open IE

Many of the current IE systems are limited in terms of scalability and portability across domains while in most corpora likes news, blog, email, encyclopedia, the extractors need to be able to extract relation tuples from across different domains. Therefore, there has been move towards next generation IE systems that can be highly scalable on large Web corpora. Etzioni et al. [10] have introduced one of the pioneering Open Information Extraction systems called TextRunner [2]. This system tackles an unbounded number of relations and eschews domain-specific training data, and scales linearly. This system does not presuppose a predefined set of relations and is targeted at all relations that can be extracted. Open IE is currently being developed in its second generation in systems such as ReVerb [12], OLLIE [12], and ClausIE [8], which extend from previous Open IE systems such as TextRunner [2], StatSnowBall [26], and WOE [23]. Figure 1 summarizes the differences of traditional IE systems and the new information extraction systems which are called Open IE [9, 10].

## 2. First Open IE generation

In the first generation, Open IE systems aimed at constructing a general model that could express a relation based on unlexicalized features such as Part-of-Speech (POS) or shallow tags e.g., a description of a verb in its surrounding context or the presence of capitalization and punctuation. While traditional IE requires relations to be specified in their input, Open IE systems use their relation-independent model as self-training to learn relations and entities in the corpora. TextRunner is one of the first Open IE systems. It applied a Naive Bayes model with POS and Chunking features that trained tuples using examples heuristically generated from the Penn Treebank. Subsequent work showed that a linear-chain Conditional Random Field (CRF) [2, 10] or Markov Logic Network [26] can be used for identifying extractable relations. Several Open IE systems have been proposed in the first generation, including TextRunner, WOE, and StatSnowBall that typically consist of the following three stages: 1) Intermediate levels of analysis and 2) Learning models and 3) Presentation, which we elaborate in the following:

### Intermediate levels of analysis

In this stage, NLP techniques such as Named Entity Recognition (NER), POS and Phrase-chunking are used. The input sequence of words are taken as input and each word in the sequence is labeled with its part of speech e.g., noun, verb, adjective by a POS tagger. A set of non overlapping phrases in the sentence is divided based on POS tags by a phrase chunked. Named entities in the sentence are located and categorized by NER. Some systems such as TextRunner, WOE used KNnext [8] work directly with the output of the syntactic and dependency parsers as shown in Figure 2. They define a method to identify useful proposition components of the parse trees. As a result, a parser will return a parsing tree including the part-of-speech of each word, the presence of phrases, grammatical structures and semantic roles for the input sentence. The structure and annotation will be essential for determining the relationship between entities for learning models of the next stage.

**Part-of-Speech**

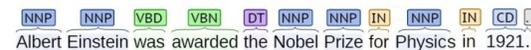

**Named Entity Recognition**

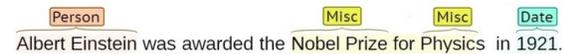

**Dependency Parsing**

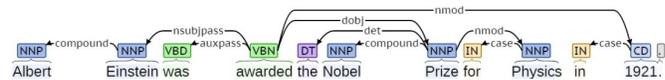

Fig. 2. POS, NER and DP analysis in the sentence "Albert Einstein was awarded the Nobel Prize for Physics in 1921".

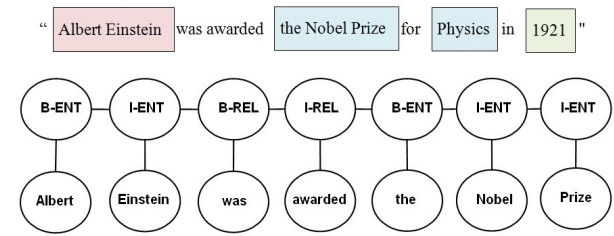

Fig. 3. A CRF is used to identify the relationship "was awarded" between "Albert Einstein" and "the Nobel Prize".

### Learning models

An Open IE would learn a general model that depicts how a relation could be expressed in a particular language. A linear-chain model such as CRF can then be applied to a sequence which is labeled with POS tags, word segments, semantic roles, named entities, and traditional forms of relation extraction from the first stage. The system will train a learning model given a set of input observations to maximize the conditional probability of a finite set of labels. TextRunner and WOE$^{pos}$ use CRFs to learn whether sequences of tokens are part of a relation. When identifying entities, the system determines a maximum number of words and their surrounding pair of entities which could be considered as possible evidence of a relation. Figure 3 shows entity pairs "Albert Einstein" and "the Nobel Prize" with the relationship "was awarded" serving to anchor the entities. On the other hand, WOE*parse* learns relations generated from corePath, a form of shortest path where a relation could exist, by computing the normalized logarithmic frequency as the probability that a relation could be found. For instance, the shortest path "Albert Einstein" ←*nsubjpass* "was awarded" →*dobj* "the Nobel Prize" presents the relationship between "Albert Einstein" and "the Nobel Prize" could be learned from the patterns "E1" ←*nsubjpass* "V" →*dobj* "E2" in the training data.

**Presentation**

In this stage, Open IE systems provide a presentation of the extracted relation triples. The sentences of the input will be presented in the form of instances of a set of relations after being labeled by the learning models. TextRunner and WOE take sentences in a corpus and quickly extract textual triples that are present in each sentence. The form of relation triples contain three textual components where the first and third denote pairs of entity arguments and the second denotes the relationship between them as (Arg1, Rel, Arg2). Figure 4 shows the differences of presentations between traditional IE and Open IE.

Additionally, with large scale and heterogeneous corpora such as the Web, Open IE systems also need to address the disambiguation of entities e.g., same entities may be referred to by a variety of names (Obama or Barack Obama or B. H. Obama) or the same string (Micheal) may refer to different entities. Open IE systems try to compute the probability that two strings denote synonymous pairs of entities based on a highly scalable and unsupervised analysis of tuples. TextRunner applies the Resolver system [25] while WOE uses the infoboxes from Wikipedia for classifying entities in the relation triples.

> Sentence: "Apple Inc. is headquartered in California"
> *Traditional IE*: Headquarters(Apple Inc., California)
> *Open IE*: (Apple Inc., is headquartered in, California)

Fig. 4. Traditional IE and Open IE extractions.

### 2.1. *Advantages and disadvantages*

Open IE systems need to be highly scalable and perform extractions on huge Web corpora such as news, blog, emails, and encyclopedias. TextRunner was tested on a collection of over 120 million Web pages and extracted over 500 million triples. This system also had a collaboration with Google on running over one billion public Web pages with noticeable precision and recall on this large-scale corpus.

First generation Open IE systems can suffer from problems such as extracting incoherent and uninformative relations. Incoherent extractions are circumstances when the system extracts relation phrases that present a meaningless interpretation of the content [2, 11]. For example, TextRunner and WOE would extract a triple such as (Peter, thought, his career as a scientist) from the sentence "*Peter thought that John began his career as a scientist*", which is clearly incoherent because "Peter" could not be taken as the first argument for relation "began" with the second argument "his career as a scientist". The second problem, uninformative extractions, occurs when Open IE systems miss critical information of a relation. Uninformative extraction is a type of error relating to light verb construction [1, 19] due to multi-word predicates being composed of a verb and an additional noun. For example, given the sentence "*Al-Qaeda claimed responsibility for the 9/11 attacks*", Open IE systems such as TextRunner return the uninformative relation (Al-Qaeda, claimed, responsibility) instead of (Al-Qaeda, claimed responsibility for, the 9/11 attack).

## 3. Second Open IE generation

In the second generation, Open IE systems focus on addressing the problem of incoherent and uninformative relations. In some cases, TextRunner and WOE do not extract the full relation between two noun phrases, and only extract a portion of the relation which is ambiguous. For instance, where it should extract the relation "is author of", it only extracts "is" as the relation in the sentence "*William Shakespeare is author of Romeo and Juliet*". Similar to first generation systems, Open IE systems in the second generation have also applied NLP techniques in the intermediate level analysis of the input and the output is processed in a similar vein to the first generation. They take a sentence as input and perform POS tagging, syntactic chunking and dependency parsing and then return a set of relation triples. However in the intermediate level analysis process, Open IE systems in the second generation focus deeply on a thorough linguistic analysis of sentences. They expose simple yet principled ways in which verbs express relationships in linguistics. Based on these linguistic relations, they obtain a significantly higher performance over previous systems in the first generation. Several Open IE systems have been proposed after TextRunner and WOE, including ReVerb, OLLIE, Christensen et al., [5], ClausIE, Vo & Bagheri [21] with two extraction paradigms, namely verb-based relation extraction and clause-based relation extraction.

### 3.1. *Verb phrase-based relation extraction*

ReVerb is one of the first systems that extracts verb phrase-based relations. This system builds a set of syntactic and lexical constraints to identify relations based on verb phrases then finds a pair of arguments for each identified relation phrase. ReVerb extracts relations by giving first priority to verbs. Then the system extracts all arguments around verb phrases that help the system to avoid common errors such as incoherent or uninformative extractions made by previous systems in the first generation. ReVerb considers three grammatical structures mediated by verbs for identifying extractable relations. In each sentence, if the phrase matches one of the three grammatical structures, it will be considered as a relation. Figure 5 depicts three grammatical structures in ReVerb. Give a sentence "Albert Einstein was awarded the Nobel Prize." for each verb V (awarded) in sentence S, it will find the longest sequence of words ($V \mid VP \mid VW^*P$) such that (1) it starts with $V$, (2) it satisfies the syntactic constraint, and (3) it satisfies the lexical constraint. As result, ($V \mid VP \mid VW^*P$) identifies "was awarded" as a relation. For each identified relation phrase $R$, it will find the nearest noun phrase $X$ to the left of $R$, which is "Albert Einstein" in this case. Then it will find the nearest noun phrase $Y$ to the right of $R$, which is "the Nobel Prize" in $S$.

---

$V \mid VP \mid VW^*P$
$V$ = verb particle? adv?
$W$ = (noun | adj | adv | pron | det)
$P$ = (prep | particle | inf. marker)

---

Fig. 5. Three grammatical structures in ReVerb [12].

Some limitations in ReVerb prevent the system from extracting all of the available information in a sentence e.g., the system could not extract the relation between "Bill Gates" and "Microsoft" in the sentence "*Microsoft co-founder Bill Gates spoke at…*" shown in Figure 6. This is due to the fact that ReVerb ignores the context of the relation by only considering verbs, which could lead to false and/or incomplete relations. Mausam et al. [14] have presented OLLIE, as an extended ReVerb system, which stands for Open Language Learning for IE. OLLIE performs deep analysis on the identified verb-phrase relation then the system extracts all relations mediated by verbs, nouns, adjectives, and others. For instance, in Figure 4 ReVerb only detects the verb-phrase to identify the relation. However, OLLIE analyzes not only the verbs but also the noun and adverb that the system could determine. As in the earlier sentence, the relations ("Bill Gates","co-founder of", "Microsoft") is extracted by OLLIE but will not be extracted using ReVerb.

OLLIE has addressed the problem in ReVerb by adding two new elements namely "*AttributedTo*" and "*ClauseModifier*" to relation tuples when extracting all relations mediated by noun, adjective, and others. "AttributeTo" is used for deterring additional information and "ClauseModifier" is used for adding conditional information as seen in sentences 2 and 3 in Figure 6. OLLIE produces high yield by extracting relations not only mediated by verbs but also mediated by nouns, and adjectives. OLLIE follows ReVerb to identify potential relations based on verb-mediated relations. The system applies bootstrapping to learn other relation patterns using its similarity relations found by ReVerb. In each pattern the system uses dependency path to connect a relation and its corresponding arguments for extracting relations mediated by noun, adjective and others. After identifying the general patterns, the system applies them to the corpus to obtain new tuples. Therefore, OLLIE extracts a higher number of relations from the same sentence compared to ReVerb.

---

1. "*Microsoft co-founder Bill Gates spoke at …* "
    OLLIE: ("Bill Gates","be co-founder of", "Microsoft")
2. "*Early astronomers believed that the earth is the center of the universe.* "
    ReVerb: ("the earth","be the center of", "the universe")
    OLLIE: ("the earth","be the center of", "the universe")
        *AttributeTo* believe; Early astronomers
3. "*If he wins five key states, Romney will be elected President.* "
    ReVerb:( "Romney", "will be elected", "President")
    OLLIE: ("Romney", "will be elected", "President")
        *ClausalModifier* if; he wins five key states

---

Fig. 6. ReVerb extraction vs. OLLIE extraction [14].

### 3.2. *Clause-based relation extraction*

A more recent Open IE system named ClausIE presented by Corro & Gemulla [8] uses clause structures to extract relations and their arguments from natural language text. Different from verb-phrase based relation extraction, this work applies clause types in sentences to separate useful pieces. ClausIE uses dependency parsing and a set of rules for domain-independent lexica to detect clauses without any requirement for training data. ClausIE exploits grammar clause structure of the English language for detecting clauses and all of its constituents in sentence. As a result, ClausIE obtains high-precision extraction of relations and also it can be flexibly customized to adapt to the underlying application domain. Another Open IE system, presented by Vo & Bagheri [21], uses clause-based approach inspired by the work presented in Corro & Gemulla [8] for open information extraction. This work proposes a reformulation of the parsing trees that will help the system identify discrete relations that are not found in ClausIE, and reduces the number of erroneous relation extractions, e.g., ClausIE incorrectly identifies 'there' as a subject of a relation in the sentence: "*In today's meeting, there were four CEOs*", which is avoided in the work by Vo and Bagheri.

Particularly, in these systems a clause can consist of different components such as subject (S), verb (V), indirect object (O), direct object (O), complement (C), and/or one or more adverbials (A). As illustrated in Table 1, a clause can be categorized into different types based on its constituent components. Both of these systems obtain and exploit clauses for relation extraction in the following manner:

**Step 1. Determining the set of clauses.** This step seeks to identify the clauses in the input sentence by obtaining the head words of all the constituents of every clause. The mapping of syntactic and dependency parsing are utilized to identify various clause constituents. Subsequently, a clause is constructed for every subject dependency, dependent constitutes of the subject, and the governor of the verb.

**Step 2. Identifying clause types.** When a clause is obtained, it needs to be associated with one of the main clause types as shown in Table 1. In lieu of the previous assertions, these systems use a decision tree to identify the different clause types. In this process, the system marks all optional adverbials after the clause types have been identified.

**Step 3. Extracting relations.** The systems extract relations from a clause based on the patterns of the clause type as illustrated in Table 1. Assuming that a pattern consists of a subject, a relation and one or more arguments, it is reasonable to presume that the most reasonable choice is to generate n-ary propositions that consist of all the constituents of the clause along with some arguments. To generate a proposition as a triple relation (Arg1, Rel, Arg2), it is essential to determine which part of each constituent would be considered as the subject, the relation and the remaining arguments. These systems identify the subject of each clause and then use it to construct the proposition. To accomplish this, they map the subject of the clause to the subject of a proposition relation. This is followed by applying the patterns of the clause types in an effort to generate propositions on this basis. For instance, for the clause type SV in Table 1, the subject presentation "Albert Einstein" of the clause is used to construct the proposition with the following potential patterns: SV, SVA, and SVAA. Dependency parsing is used to forge a connection between the different parts of the pattern. As a final step, n-ary facts are extracted by placing the subject first followed by the verb or the verb with its constituents. This is followed by the extraction of all the constituents following the verb in the order in which they appear. As a result, these systems link all arguments in the propositions in order to extract triple relations.

Table 1. Clause types [8, 16]; S: Subject, V: Verb, A: Adverbial, C: Complement, O: Object

| Clause types | Sentences | Patterns | Derived clauses |
|---|---|---|---|
| SV | Albert Einstein died in Princeton in 1955. | SV<br>SVA<br>SVA<br>SVAA | (Albert Einstein, died)<br>(Albert Einstein, died in, Princeton)<br>(Albert Einstein, died in, 1955)<br>(Albert Einstein, died in, 1955, [in] Princeton) |
| SVA | Albert Einstein remained in Princeton until his death. | SVA<br>SVAA | (Albert Einstein, remained in, Princeton)<br>(Albert Einstein, remained in, Princeton, until his death) |
| SVC | Albert Einstein is a scientist of the 20th century. | SVC<br>SVCA | (Albert Einstein, is, a scientist)<br>(Albert Einstein, is, a scientist, of the 20 the century) |
| SVO | Albert Einstein has won the Nobel Prize in 1921. | SVO<br>SVOA | (Albert Einstein, has won, the Nobel Prize)<br>(Albert Einstein, has won, the Nobel Prize, in 1921) |
| SVOO | RSAS gave Albert Einstein the Nobel Prize. | SVOO | (RSAS, gave, Albert Einstein, the Nobel Prize) |
| SVOA | The doorman showed Albert Einstein to his office. | SVOA | (The doorman, showed, Albert Einstein, to his office) |
| SVOC | Albert Einstein declared the meeting open. | SVOC | (Albert Einstein, declared, the meeting, open) |

### 3.3. *Advantages and disadvantages*

The key differentiating characteristic of these systems is a linguistic analysis that guides the design of the constraints in ReVerb and features analysis in OLLIE. These systems address incoherent and uninformative extractions which occur in the first generation by identifying a more meaningful relation phrase. OLLIE expands the syntactic scope of Reverb by identifying relations mediated by nouns and adjectives around verb phrase. Both ReVerb and OLLIE outperform the previous systems in the first Open IE generation. Another approach in the second generation, clause-based relation extraction, uses dependency parsing and a set of rules for domain-independent lexica to detect clauses for extracting relations without raining data. They exploit grammar clauses of the English language to detect clauses and all of their constituents in a sentence. As a result, systems such as ClausIE obtain high-precision extractions and can also be flexibly customized to adapt to the underlying application domain.

In the second Open IE generation, binary extractions have been identified in ReVerb and OLLIE, but not all relationships are binary. Events can have time and location and may take several arguments (e.g., "*Albert Einstein was awarded the Nobel Prize for Physics in 1921.*"). It would be essential to extend Open IE to handle n-ary and even nested extractions.

## 4. Application Areas

There are several areas where Open IE systems can be applied:

First, the ultimate objectives of Open IE systems are to enable the extraction of knowledge that can be represented in structured form and in human readable format. The extracted knowledge can be then used to answer questions [17, 21]. For instance, TextRunner can support user input queries such as "(?, kill, bacteria)" or "(Barack Obama, ?, U.S)" similar to Question Answering systems. By replacing the question mark in the triple, questions such as "what kills bacteria" and "what are the relationships between Barack Obama and U.S" will be developed and can be answered.

Second, Open IE could be integrated and applied in many higher levels of NLP tasks such as text similarity or text summarization [6, 7, 13]. Relation tuples from Open IE systems could be used to infer or measure the redundancy between sentences based on the facts extracted from the input corpora.

Finally, Open IE can enable the automated learning and population of an upper level ontology due to its ability in the scalable extraction of information across domains [18]. For instance, Open IE systems can enable the learning of a new biomedical ontology by automatically reading and processing published papers in the literature on this topic.